# RDD2022: A multi-national image dataset for automatic Road Damage Detection

Deeksha Arya[1,2,*], Hiroya Maeda[3], Sanjay Kumar Ghosh[1,4], Durga Toshniwal[1,5], Yoshihide Sekimoto[2]

[1]Centre for Transportation Systems (CTRANS), Indian Institute of Technology Roorkee, Roorkee, India
[2]Centre for Spatial Information Science, The University of Tokyo, Tokyo, Japan
[3]UrbanX Technologies, Inc., Tokyo, Japan
[4]Department of Civil Engineering, Indian Institute of Technology Roorkee, India
[5]Department of Computer Science and Engineering, Indian Institute of Technology Roorkee, India

## Abstract

The data article describes the Road Damage Dataset, RDD2022, which comprises 47,420 road images from six countries, Japan, India, the Czech Republic, Norway, the United States, and China. The images have been annotated with more than 55,000 instances of road damage. Four types of road damage, namely longitudinal cracks, transverse cracks, alligator cracks, and potholes, are captured in the dataset. The annotated dataset is envisioned for developing deep learning-based methods to detect and classify road damage automatically. The dataset has been released as a part of the Crowd sensing-based Road Damage Detection Challenge (CRDDC'2022). The challenge CRDDC'2022 invites researchers from across the globe to propose solutions for automatic road damage detection in multiple countries. The municipalities and road agencies may utilize the RDD2022 dataset, and the models trained using RDD2022 for low-cost automatic monitoring of road conditions. Further, computer vision and machine learning researchers may use the dataset to benchmark the performance of different algorithms for other image-based applications of the same type (classification, object detection, etc.).

## Background and Summary

The Road Damage Dataset, RDD2022, is an extended version of the existing RDD2020 [1,2] dataset. Fig. 1 shows the evolution of the dataset over the years. The corresponding statistics are compared in Fig. 2. Firstly, the dataset RDD2018 [3] was introduced in 2018, comprising 9,053 road images with 15,435 road damage instances. The RDD2018 dataset captured the information on eight types of road damage (Table 1) and was utilized for Road Damage Detection Challenge in 2018, organized as an IEEE Big Data Cup. Fifty-nine teams from 14 countries participated in the challenge and proposed multiple solutions for automatic road damage detection using RDD2018. The participants also suggested some improvements in the annotation files of the RDD2018 dataset.

Considering these suggestions and the imbalance in the road damage categories captured in the RDD2018 dataset, its authors introduced the extended version of the dataset, RDD2019[4]. RDD2019 was prepared by correcting the annotations in the RDD2018 dataset and augmenting it using the Generative Adversarial Network (GAN). It included 13,133 images with 30989 road damage instances. The expanded dataset helped achieve better accuracy for road damage detection and classification models.
In 2020, Arya et al. (2020) [5,6] attempted to apply the models trained using RDD2019 to detect road damage outside Japan. Experiments were performed using the data from India and the Czech Republic. The outcome revealed that the performance of the models trained using RDD2019 degraded significantly when applied to roads outside Japan. Consequently, the authors proposed another dataset, RDD2020. The RDD2020[1] dataset was prepared by combining the RDD2019[4] dataset with newly collected data from India and the Czech Republic. Along with increasing the number of images and damage instances to form RDD2020, the types of road damage considered were also updated.

* Corresponding author: deeksha@ct.iitr.ac.in

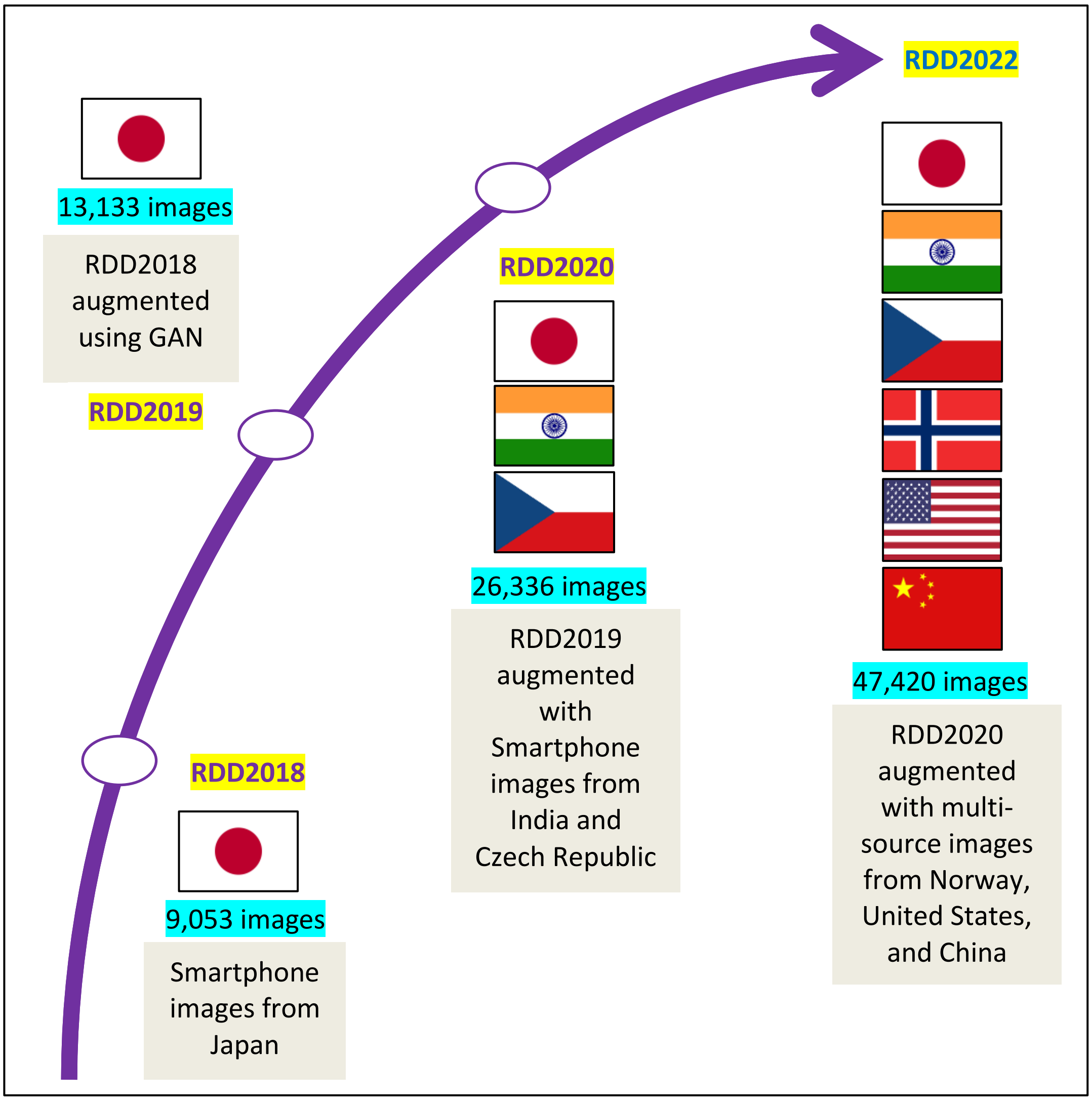

Figure 1: Schematic overview of the study design: Evolution of Road Damage Datasets from RDD2018 to the proposed dataset RDD2022

Since RDD2018 and RDD2019 captured the data from a single country, the damage categories defined in the Japanese Road Maintenance and Repair Guidebook 2013[7] were used. However, in RDD2020, the involvement of multiple countries required considering multiple road damage standards. Since the criteria for assessing Road Marking deteriorations such as Crosswalk or White Line Blur vary significantly across different countries, these categories were excluded from the RDD2020 dataset [1,5]. Subsequently, the following four damage categories – Longitudinal Cracks (D00), Transverse Cracks (D10), Alligator Cracks (D20), and Potholes (D40), were included in the RDD2020 dataset.

The RDD2020 dataset was utilized for organizing the Global Road Damage Detection Challenge (GRDDC'2020) [8]. The challenge invited researchers across the globe to propose a single model for monitoring road conditions in the three countries (India, Japan, and the Czech Republic). One hundred twenty-one teams participated in the challenge and proposed multiple solutions with varying accuracy and resource requirements. The best performing model utilized the YOLOv5-based ensemble model and achieved an F1 score of 0.67.

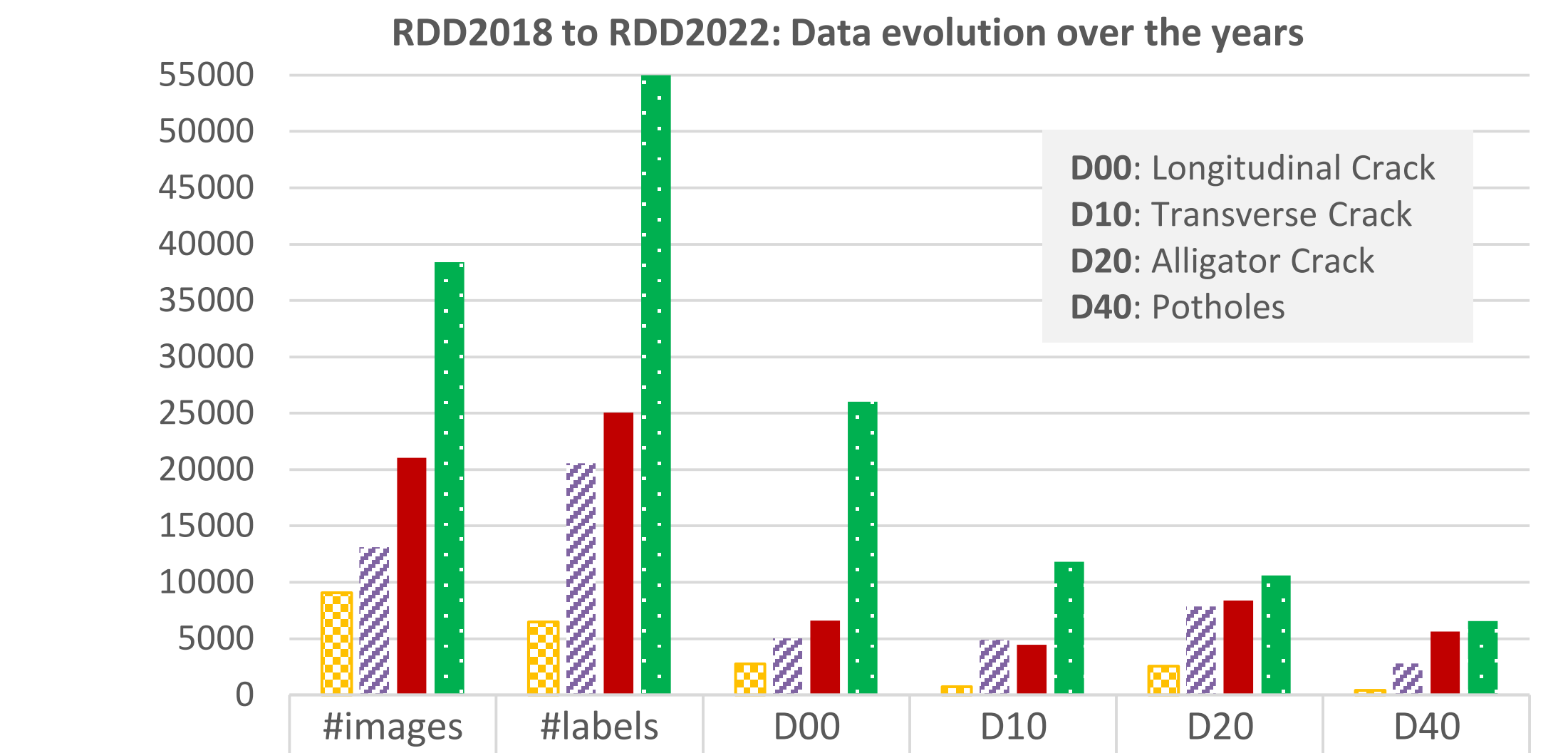

| | #images | #labels | D00 | D10 | D20 | D40 |
|---|---|---|---|---|---|---|
| RDD2018 | 9053 | 6460 | 2768 | 742 | 2541 | 409 |
| RDD2019 | 13133 | 20561 | 5006 | 4911 | 7846 | 2798 |
| RDD2020* | 21041 | 25046 | 6592 | 4446 | 8381 | 5627 |
| RDD2022* | 38385 | 55007 | 26016 | 11830 | 10617 | 6544 |

*Statistics listed only for the publically available training set of the RDD2020 and RDD2022

Figure 2: Statistical comparison of the road damage datasets from 2018 to 2022

Table 1: Road damage types in the RDD2018 dataset [3] and their definitions

| ***Damage Type*** | | | ***Detail*** | ***Class Name*** |
|---|---|---|---|---|
| Crack | Linear Crack | Longitudinal | Wheel mark part | D00 |
| | | | Construction joint part | D01 |
| | | Lateral | Equal interval | D10 |
| | | | Construction joint part | D11 |
| | Alligator Crack | | Partial pavement, overall pavement | D20 |
| Other Corruption | | | Rutting, bump, pothole, separation | D40 |
| | | | Crosswalk blur | D43 |
| | | | White line blur | D44 |

Further, the analysis conducted by Arya et al. [6] indicated that adding data from other countries helps improve the generalizability of models trained for detecting road damage in any country. This analysis and the tremendous success of the GRDDC'2020 [8] are the motivation behind introducing the current dataset RDD2022. It aims to solve road damage detection for a more extensive set of countries, targeting solutions for India, Japan, the Czech Republic, Norway, China, and the United States. The sample images for the four types of damage considered in RDD2022 are shown in fig. 3. The country-wise samples are included in the Data Records section of the manuscript.

RDD2022 is prepared and released as a part of the Crowd sensing-based Road Damage Detection Challenge (CRDDC'2022 - https://crddc2022.sekilab.global/ ). It is divided into two sets: train and test. The images in the train set are released with annotations. However, the images in the test set are for testing the models proposed by the CRDDC participants and hence have been released without annotations. The country-wise distribution of images and labels for RDD2022 is presented in Fig. 4. The distribution of different labels (damage categories) in the train set is summarized in Fig. 5.

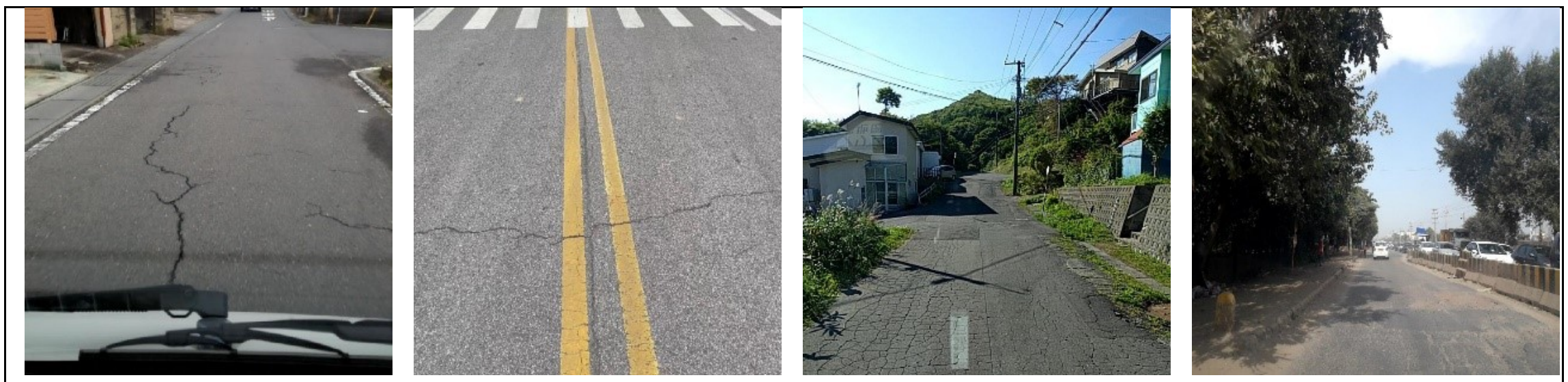

a. b. c. d.

Figure 3: Sample images for road damage categories considered in the data. a. Longitudinal Crack (D00), b. Transverse Crack (D10), c. Alligator Crack (D20), d. Pothole (D40).

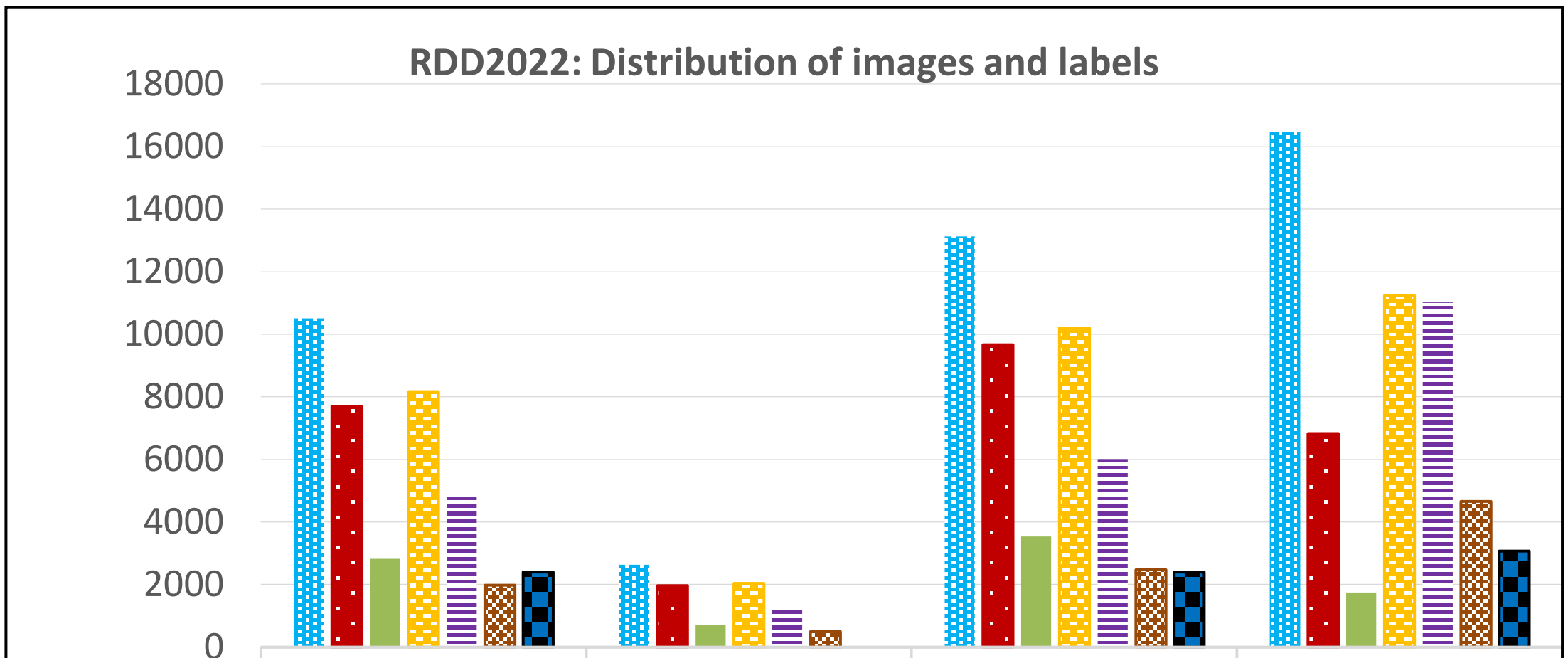

| | #train_images | #test_images | #total_images | #train_labels |
|---|---|---|---|---|
| Japan | 10506 | 2627 | 13133 | 16470 |
| India | 7706 | 1959 | 9665 | 6831 |
| Czech | 2829 | 709 | 3538 | 1745 |
| Norway | 8161 | 2040 | 10201 | 11229 |
| US | 4805 | 1200 | 6005 | 11014 |
| China_M | 1977 | 500 | 2477 | 4650 |
| China_D | 2401 | 0 | 2401 | 3068 |

Figure 4: Data statistics for RDD2022: Country-wise distribution of images and labels

Further details regarding the dataset are provided as follows. For data included from Japan, India, and the Czech Republic, the readers may refer to the article [6]. A summary of the locations covered in these countries is included here to comprehensively describe the heterogeneity and robustness considered in the proposed RDD2022 dataset.

i. **Japan**: The data is collected from seven local governments in Japan (Ichihara city, Chiba city, Sumida ward, Nagakute city, Adachi city, Muroran city, and Numazu city). The municipalities have snowy and urban areas that vary widely from the perspective of regional characteristics like weather and budgetary constraints.

ii. **India**: The data from India includes images captured from local roads, state highways, and national highways, covering the metropolitan (Delhi, Gurugram) as well as non-metropolitan regions (mainly from the state of Haryana). All these images have been collected from plain areas. Road selection and time of data collection were decided based on road accessibility, atmospheric conditions, and traffic volume.

iii. **Czech Republic:** A substantial portion of road images was collected in Olomouc, Prague, and Bratislava municipalities and covered a mix of first-class, second-class, and third-class roads and local roads. A smaller portion of the road image dataset was collected along D1, D2, and D46 motorways to enhance the resilience of the targeted model.

iv. **Norway:** The data from Norway consists of two classes of roads a) Expressways and b) County Roads (or Low Volume Roads). Both types of road classes are asphalt pavements. Data collection is done by the Norwegian Public Road Administration (Statens Vegvesen, SVV) and Inlandet Fylkeskommune (IFK). Images provided by SVV were collected on European Route E14, connecting the city of Trondheim in Norway to Sundsvall in Sweden. At the same time, the images from IFK belong to different county roads within Inllanndet County in Norway. The images were collected without any control over daytime/light, and all the images are natural without further processing. Further, the dataset captures diverse backgrounds, including clear grass fields, snow-covered areas, and conditions after rain. Furthermore, images with different illuminances, such as high sunlight and overcast weather resulting in daylight, are considered.

v. **United States** – The data from the United States consists of Google Street View images covering multiple locations, including California, Massachusetts, and New York. The image count for each site is provided in fig. 6.

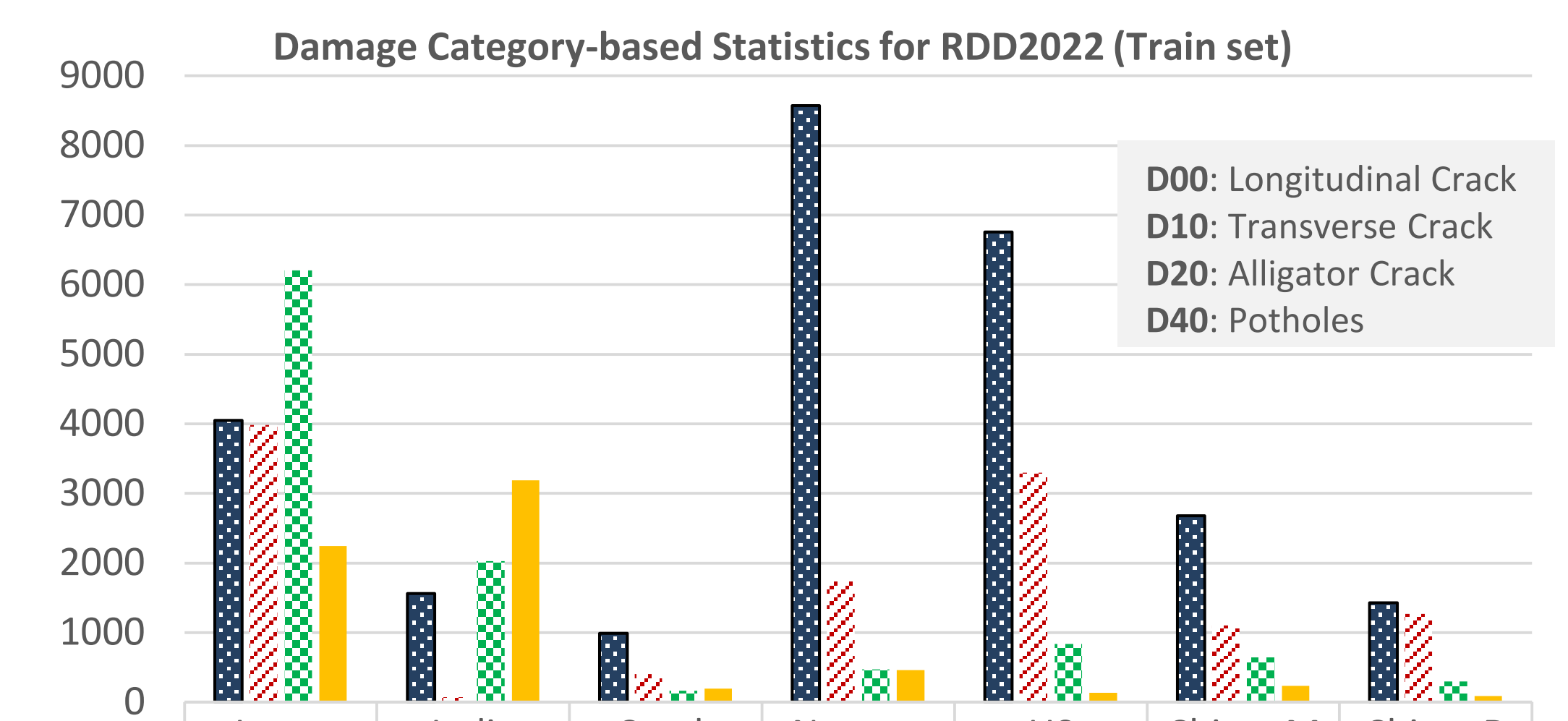

| | Japan | India | Czech | Norway | US | China_M | China_D |
|---|---|---|---|---|---|---|---|
| D00 | 4049 | 1555 | 988 | 8570 | 6750 | 2678 | 1426 |
| D10 | 3979 | 68 | 399 | 1730 | 3295 | 1096 | 1263 |
| D20 | 6199 | 2021 | 161 | 468 | 834 | 641 | 293 |
| D40 | 2243 | 3187 | 197 | 461 | 135 | 235 | 86 |

Figure 5: Damage Category-based data statistics for RDD2022

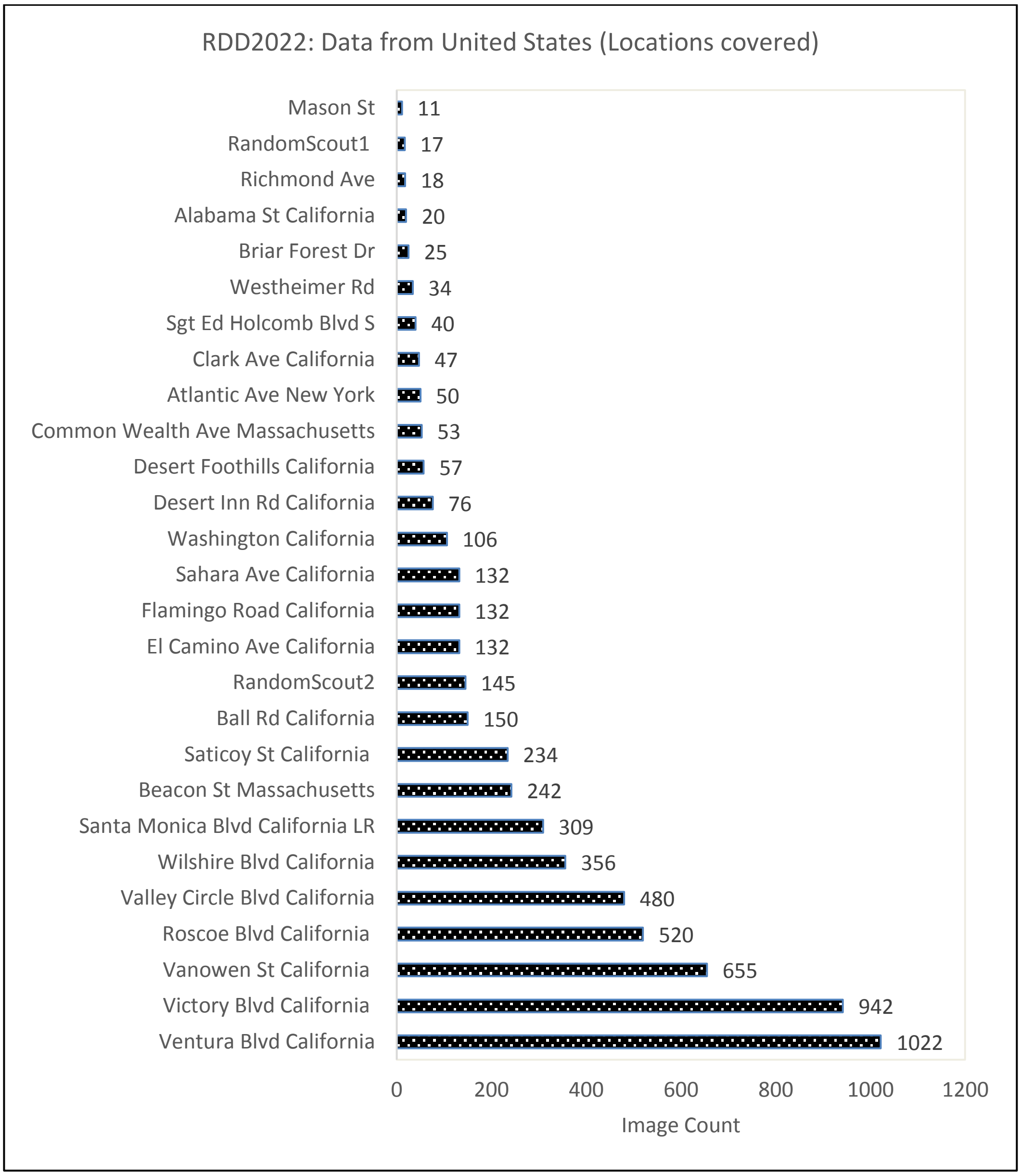

Figure 6: Details of the locations covered from the United States in RDD2022 data

vi. **China:** RDD2022 considers two types of data from China: (a) images captured by Drones (represented as China_Drone or China_D), and (b) the images captured using Smartphone-mounted MotorBikes (represented as China_MotorBike or China_M). The drone images were obtained from Dongji Avenue in Nanjing, China. The MotorBike images were collected on Jiu long hu campus, Southeast University. Images with normal light, under a shadow environment, and wet stains are covered.

Asphalt concrete pavement is considered with a few exceptions in all six subsets of RDD2022. The methods used to capture the images and generate the annotations in RDD2022 are discussed in the following section. A comparative summary of RDD2022 with its previous versions is presented in Table 2, showing the quantum of improvement in the proposed dataset.

Table 2: Comparative summary of the proposed RDD2022 dataset with previous versions

| | RDD2018[3] | RDD2019[4] | RDD2020[1] | RDD2022 (proposed) |
|---|---|---|---|---|
| **#Images** | 9,053 | 13,133 | 26,336 | 47,420 |
| **Image capturing device** | Smartphones | Smartphones | Smartphones | Smartphones, High-resolution Cameras, Google Street View images |
| **Vehicle for data collection** | Cars | Cars | Cars | Cars, Motorbikes, Drones |
| **Image resolution** | 600x600 | 600x600 | 600x600, 720x720 | 512x512, 600x600, 720x720, 3650x2044 |
| **Road view captured in the images** | Wide view (Road surface and surroundings captured horizontally) | Wide view | Wide view | Wide view, extra-wide view (using two cameras in Norway), top-down view |
| **#Damage categories** | 8 | 9 | 4 | 4 |
| **#Labels** | 15,435 (train + test) for 8 categories | 30,989 (train + test) for 9 categories | 21,041 (Released for training) for 4 categories | 55,007 (Released for training) for 4 categories |
| **Annotation method** | LabelImg tool | LabelImg tool | LabelImg tool | LabelImg tool and Computer Vision Annotation Tool (CVAT) |
| **#Countries covered** | 1 (Japan) | 1 (Japan) | 3 (Japan, India, Czech Republic) | 6 (Japan, India, Czech Republic, Norway, United States, China) |
| **Geographical diversity** | High (7 municipalities with diverse regional and weather characteristics) | Same as RDD2018 | Higher (internationally diverse regions are considered along with a local variation for each country) | Highest (international diversity further enhanced) |

## Methods

The RDD2022 data has been prepared using two steps: image acquisition and damage annotation. For each of the six countries specified above, road images are captured and labelled using software to mark the road damage captured in the images. The details of these two processes are provided as follows.

**Image Acquisition**

The images in RDD2022 have been acquired using different methods for different countries. For India, Japan, and the Czech Republic, smartphone-mounted vehicles (cars) were utilized to capture road images. The installation setup of the smartphone in the car is shown in fig. 7. In some cases, the setup with the smartphone mounted on the windshield (inside the car) was also used. Images of resolution 600x600 are captured for Japan and the Czech Republic. For India, images are captured at a resolution of 960x720 and later resized to 720x720 to maintain uniformity with the data from Japan and the Czech Republic.

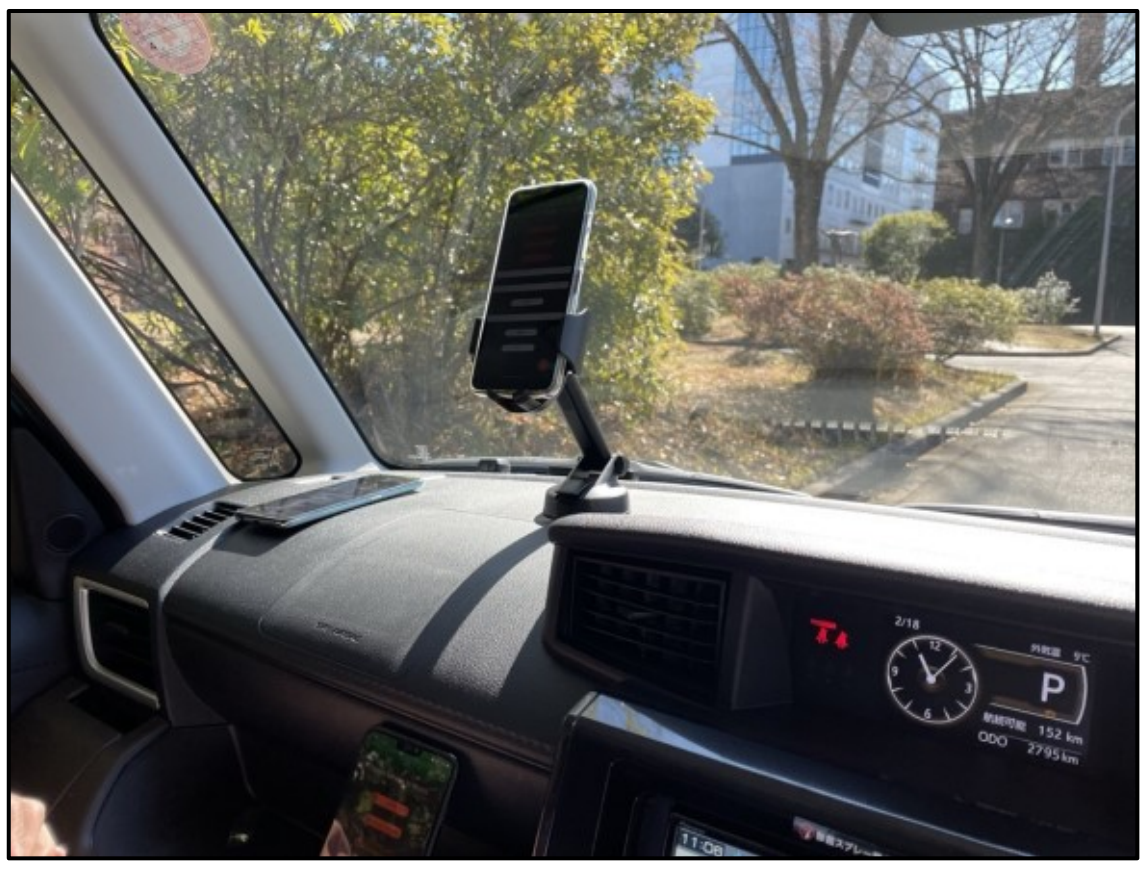

Figure 7: Sample Installation Setup of the smartphone in the car: Image acquisition for data from India, Japan, and the Czech Republic

For Norway, instead of smartphones, high-resolution cameras mounted inside the windshield of a specialized vehicle, ViaPPS, were used for data collection (Fig. 8). ViaPPS System employs two Basler_Ace2040gc cameras (Fig. 9) with Complementary metal oxide semiconductor (CMOS) sensor to capture images and then stitches them into one wide view image of a typical resolution 3650x2044.

In contrast, the data for the United States comprises Google Street View Images (vehicle-based) of the resolution 640x640. Likewise, for China, two types of image-acquisition methods are considered. The first method includes a camera mounted on motorbikes moving at an average speed of 30km/h; the corresponding dataset is referred to as China_MotorBike or China_M. The second method uses a six-motor UAV manufactured by DJI (M600 Pro) for pavement image collection, resulting in China_Drone (or China_D) data. A controllable three-axis gimbal was mounted at the bottom of the UAV to hold the camera and allow 360-degree rotation for capturing the China_Drone data. The corresponding setup is shown in fig. 10. The resolution of images for the data from China is 512x512.

It may be noted that the China_Drone data has been included only in the training set of the proposed RDD2022 data to enhance the heterogeneity. The main goal of RDD2022 data still aligns with that of RDD2020, which focuses on low-cost affordable automatic road damage detection considering feasible methods for the public.

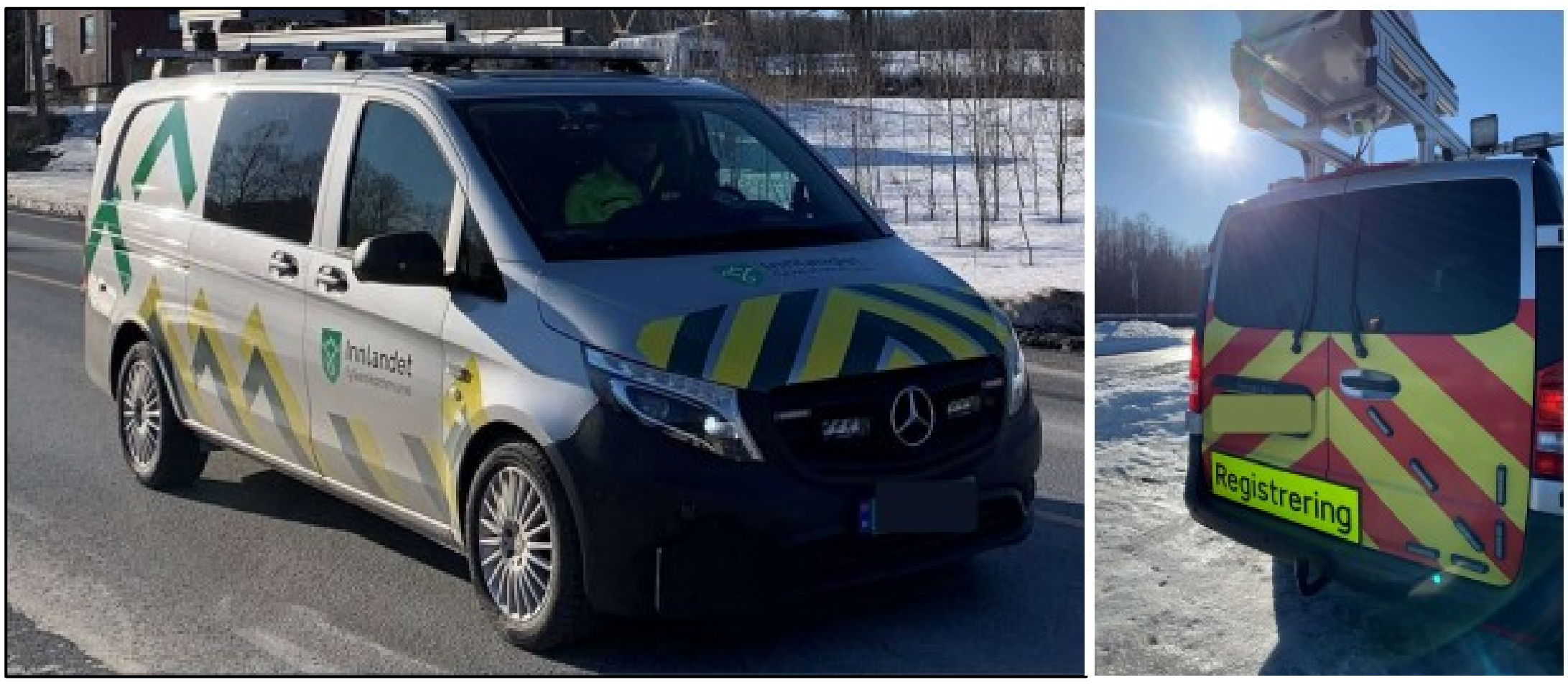

Figure 8: ViaPPS vehicle used to collect data from Norway

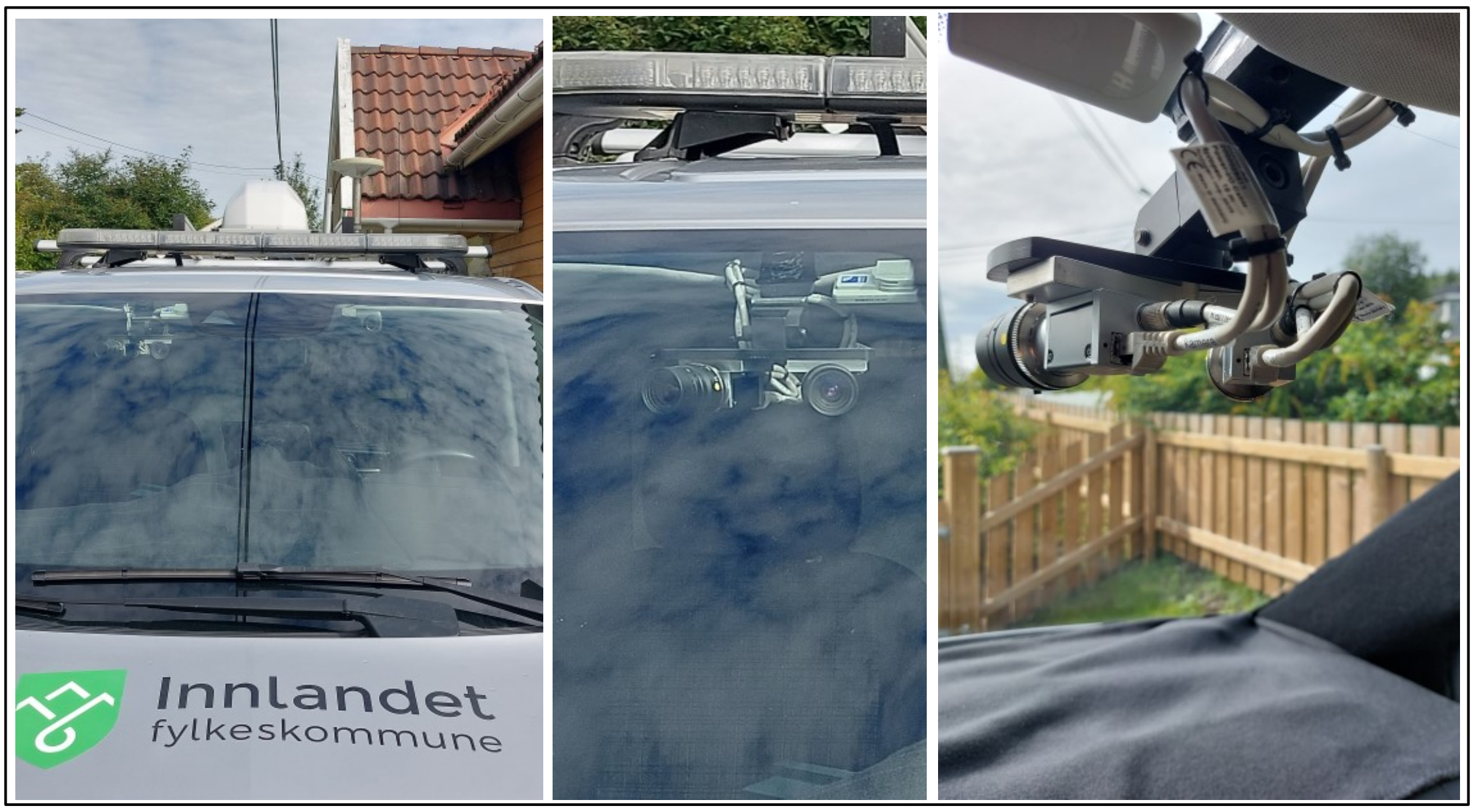

Figure 9: Camera set-up in the ViaPPS vehicle used to collect data from Norway

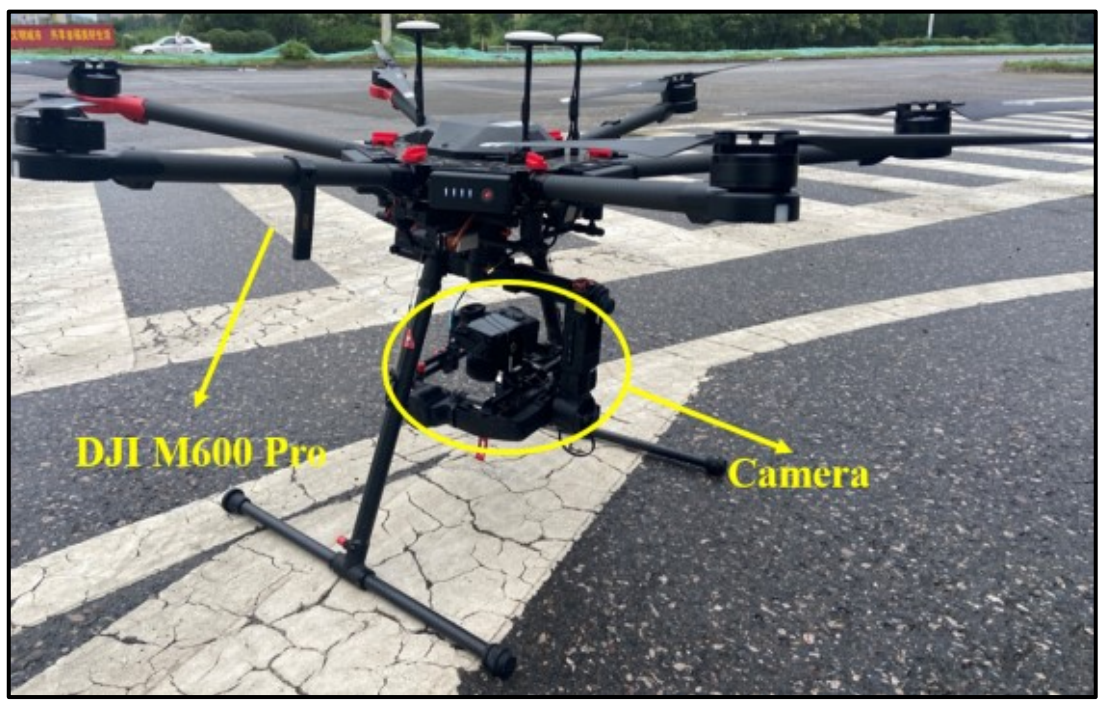

Figure 10: The drone and camera set-up used to capture China_Drone data included in RDD2022

### Annotation

RDD2022 includes annotation for road damage present in the image. The software LabelImg has been used to annotate the images except for the data from Norway. For Norway, another software Computer Vision Annotation Tool (CVAT), was utilized. Both the software packages are open source through public repositories: https://github.com/heartexlabs/labelImg and https://github.com/opencv/cvat, respectively.

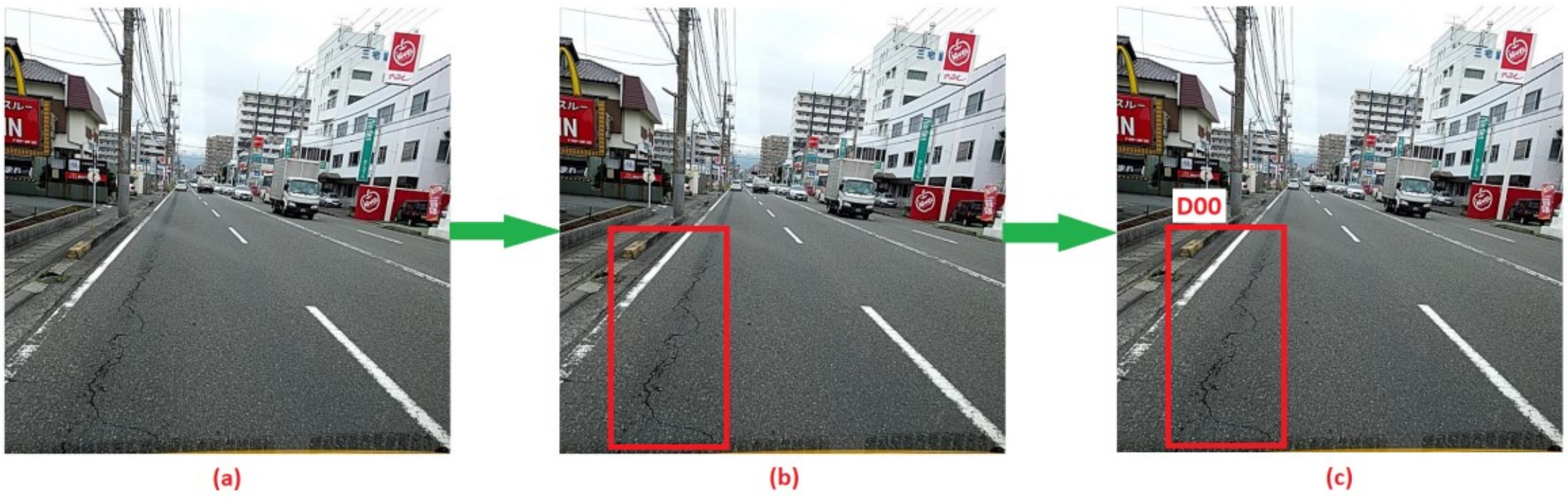

Figure 11: Annotation Pipeline (a) input image, (b) image with bounding boxes, (c) final annotated image containing bounding boxes and class label (D00 in this case)

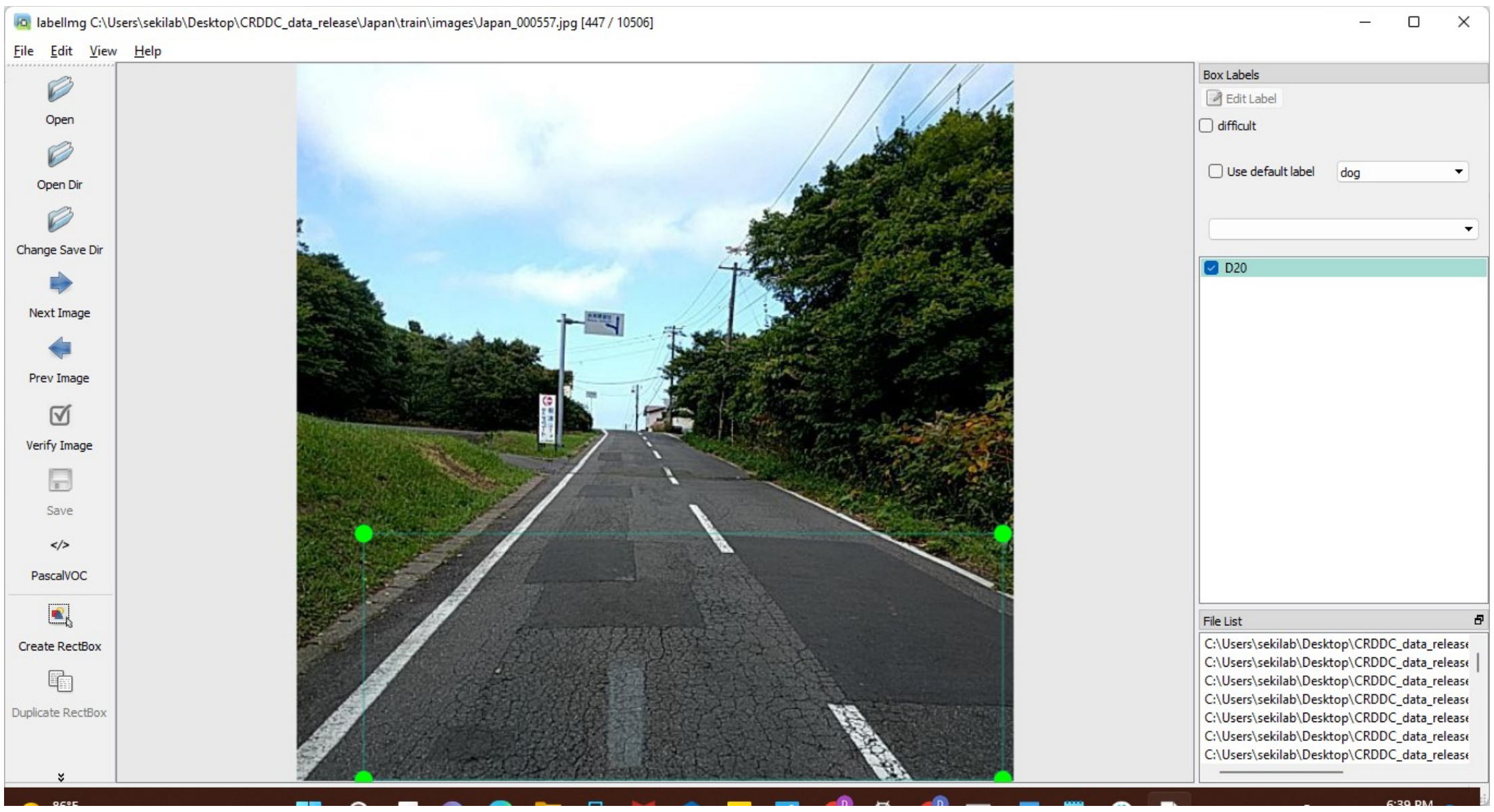

Figure 12: Sample Annotation in LabelImg tool.

The annotation pipeline is the same as the one used for RDD2020[1] and is presented in Fig. 11. Sample annotation in the LabelImg tool is shown in Fig. 12. All recognized damage instances were annotated by enclosing them with bounding boxes and classified by attaching the proper class label. Class labels and bounding box coordinates, defined by four decimal numbers (xmin, ymin, xmax, ymax), were stored in the XML format similar to PASCAL VOC [9]. The sample annotation file is shown in Fig. 13.

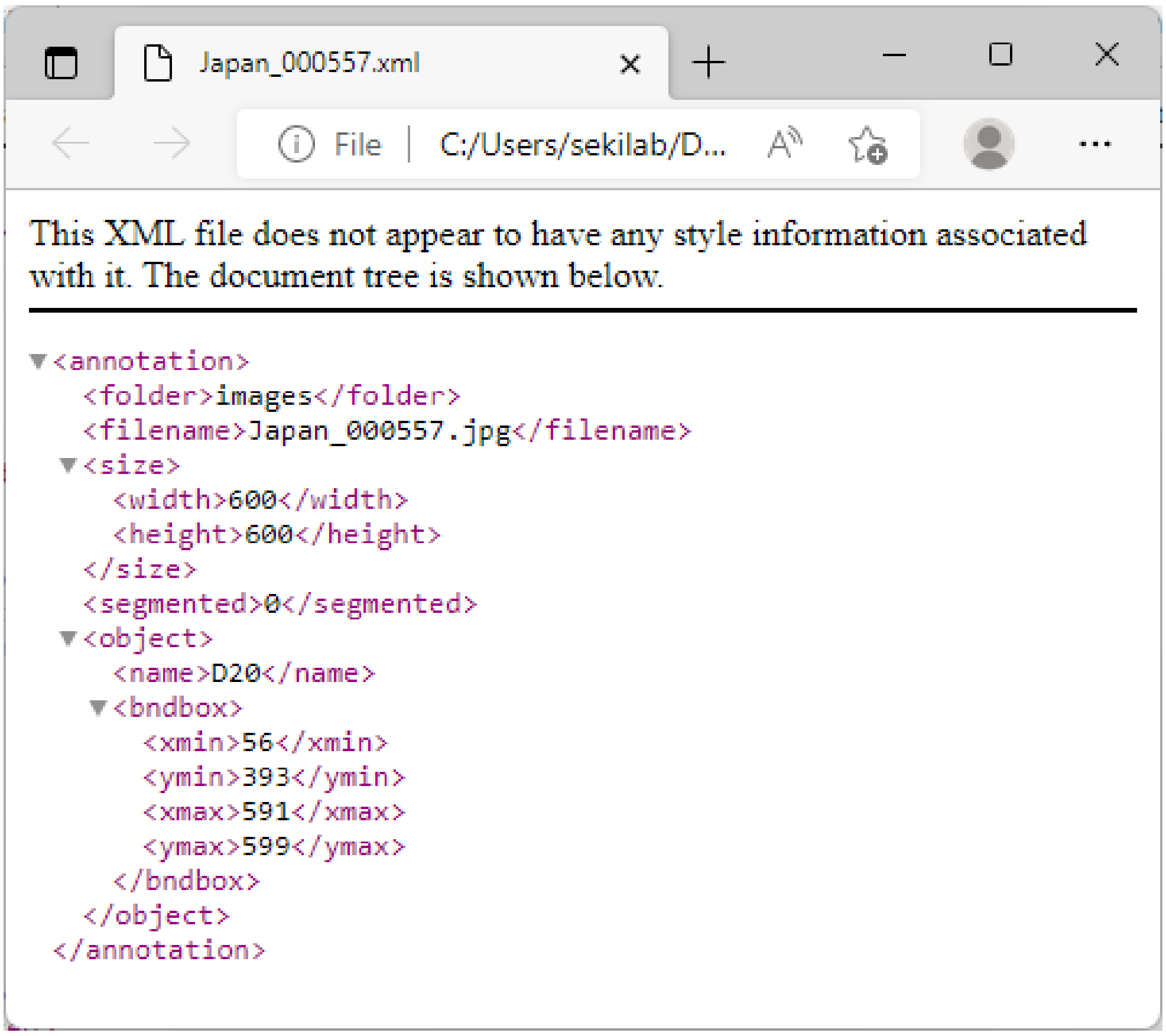

Figure 13: Sample annotation file: XML file corresponding to annotation performed in Fig. 11

## Data Records

The RDD2022 data can be accessed at the GitHub repository: https://github.com/sekilab/RoadDamageDetector. Figure 14 shows the corresponding directory structure. It comprises seven folders, described as follows:

1. China_Drone: It contains the data from China collected using Drones. Sample images are shown in Fig. 15.
2. China_MotorBike: It covers the data from China collected using Motorbike. Fig. 16 shows the sample images.
3. Czech: It includes the Czech Republic data collected using vehicle-mounted Smartphones. The sample images are shown in Fig. 17.
4. India: It consists of the data from India collected using vehicle-mounted Smartphones. Fig. 18 shows the sample images.
5. Japan: It includes the data from Japan, collected using vehicle-mounted Smartphones. Fig. 19 shows the sample images.
6. Norway: It includes the Norway data collected using vehicle-mounted high-resolution cameras. Fig. 20 shows the sample images.
7. United_States: It contains the United States data collected using Google Street View. Fig. 21 shows the sample images.

Each folder contains a sub-folder, “train,” which includes images (.jpg) and their annotations (.xml), as described in the previous section. The corresponding statistics (number of images, number of labels, etc.) are presented in Fig. 4 and Fig. 5. The sample annotation file (.xml) is shown in Fig. 13.

Along with the “train,” a sub-folder “test” is also included in each of the folders, except for China_Drone. The “test” directory contains images (.jpg files) for testing the models trained using “train” data. The annotations for “test” images have not been released. The users may utilize the leader boards on the CRDDC’2022 website to assess the prediction of their models for images in “test” data.

```
Select Anaconda Prompt

(base) C:\Users\sekilab\Deeksha\CRDDC_2022_Data_release>tree
Folder PATH listing for volume Windows
Volume serial number is 520B-286B
C:.
└───RDD2022
    ├───China_Drone
    │   └───train
    │       ├───annotations
    │       │   └───xmls
    │       └───images
    ├───China_MotorBike
    │   ├───test
    │   │   └───images
    │   └───train
    │       ├───annotations
    │       │   └───xmls
    │       └───images
    ├───Czech
    │   ├───test
    │   │   └───images
    │   └───train
    │       ├───annotations
    │       │   └───xmls
    │       └───images
    ├───India
    │   ├───test
    │   │   └───images
    │   └───train
    │       ├───annotations
    │       │   └───xmls
    │       └───images
    ├───Japan
    │   ├───test
    │   │   └───images
    │   └───train
    │       ├───annotations
    │       │   └───xmls
    │       └───images
    ├───Norway
    │   ├───test
    │   │   └───images
    │   └───train
    │       ├───annotations
    │       │   └───xmls
    │       └───images
    └───United_States
        ├───test
        │   └───images
        └───train
            ├───annotations
            │   └───xmls
            └───images
```

Figure 14: The directory structure for the RDD2022 dataset

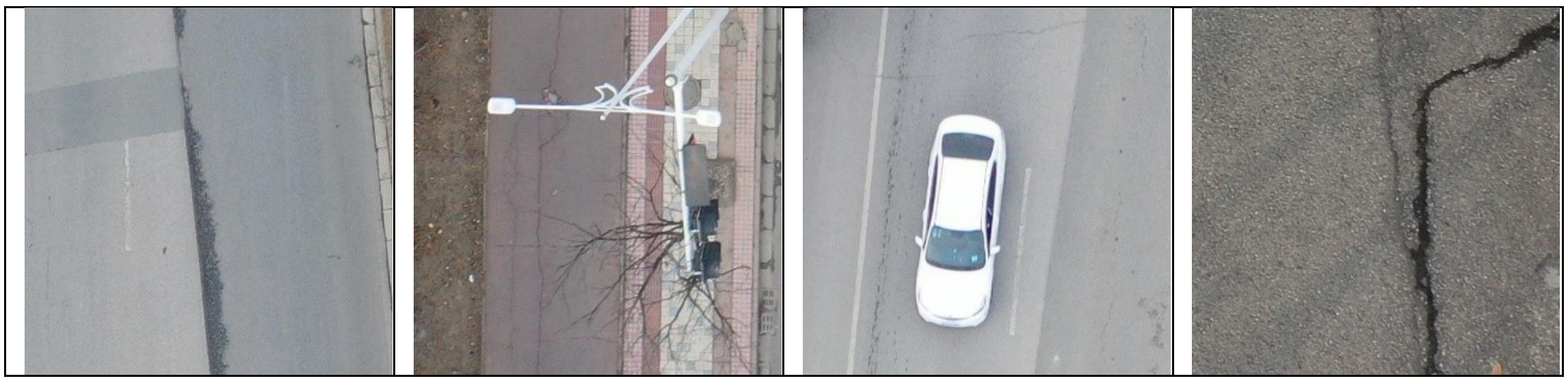

Figure 15: Sample images from China: Collected using Drones

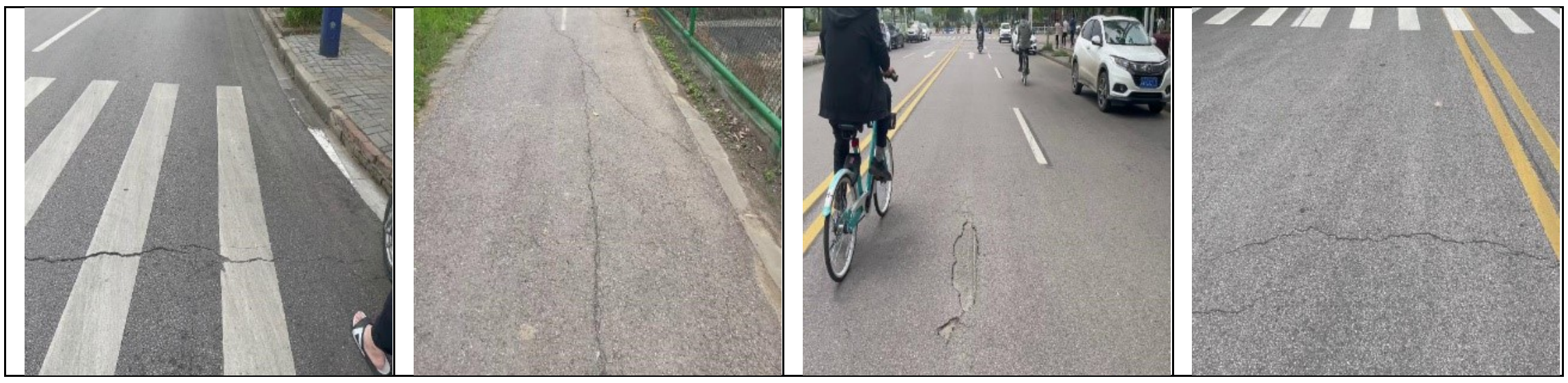

Figure 16: Sample images from China: Collected using MotorBikes

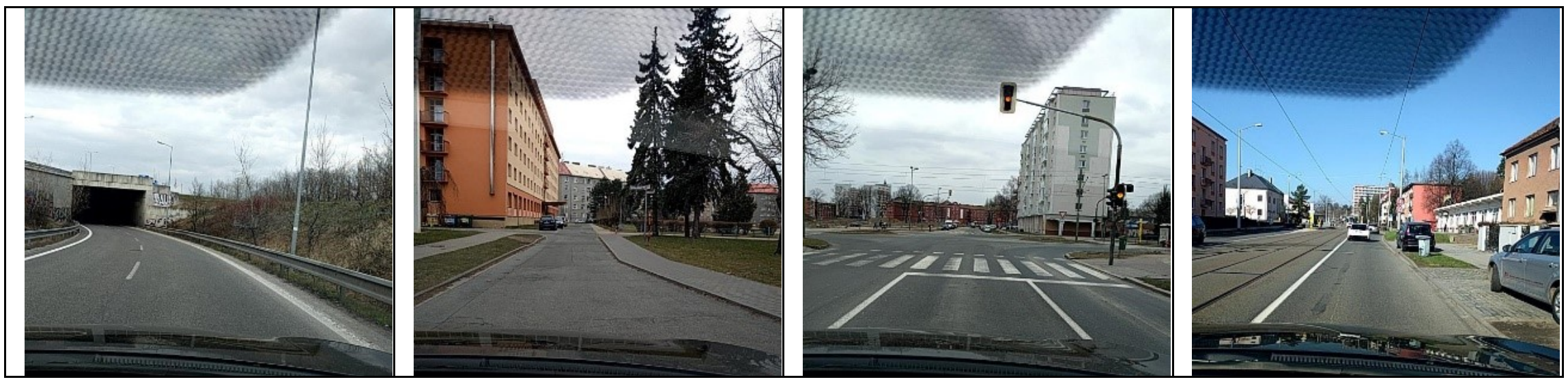

Figure 17: Sample images from the Czech Republic

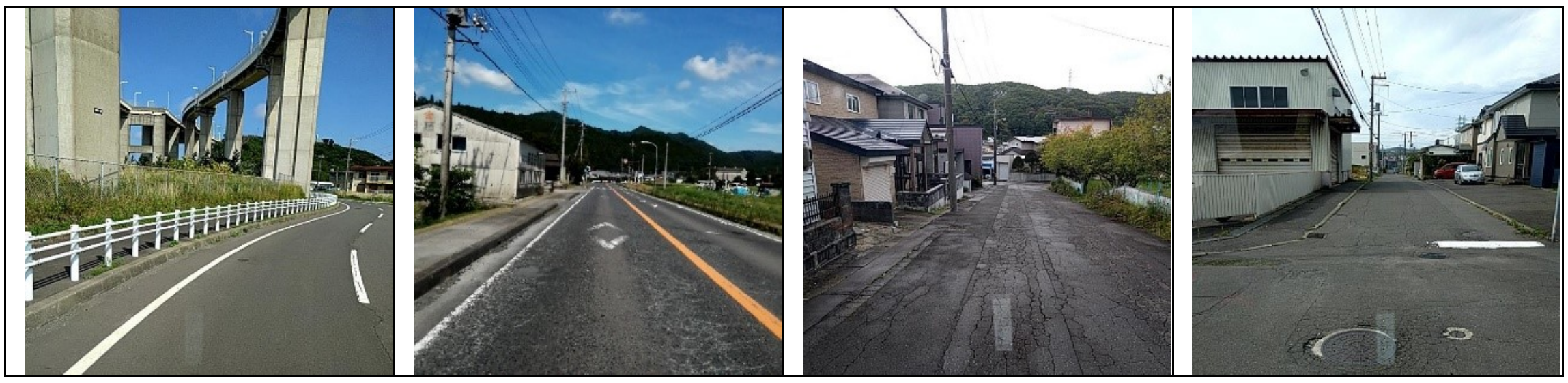

Figure 18: Sample images from Japan

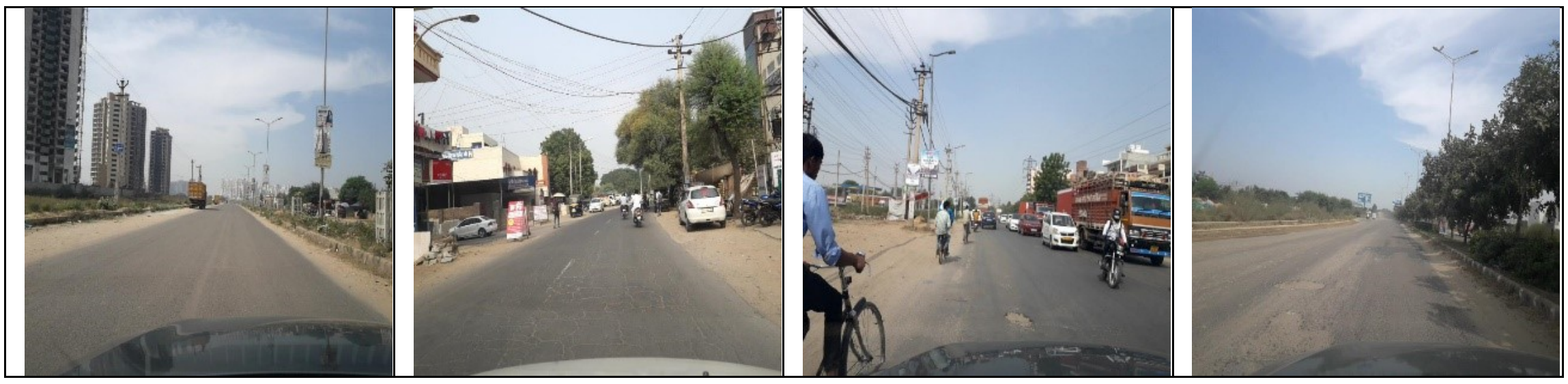

Figure 19: Sample images from India

## Usage Notes

The RDD2022 dataset is prepared following the format of the famous PASCAL Visual Object Classes (VOC) datasets [9]; hence, many existing methods used in the field of image processing may be applied to it. The usage of RDD2022 may be considered in the following two types of scenarios:

(i) Images and Annotations - when users want to use both the images and annotations included in RDD2022 data: The users may wish to use the RDD2022 data directly (as it is) or augment it for extended applications. A few applications are listed below.
   a. Direct use
      i. The RDD2022 data lays down the foundation for smartphone-based automatic road damage detection and is valuable for municipalities and road organizations for low-cost inspection of road conditions.

ii. The data may be used to develop new deep convolutional neural network architectures or modify the existing architectures to improve the network’s performance.
iii. Researchers may utilize the data to train, validate, and test the algorithms for automatically identifying road damages in six countries.
iv. The images in RDD2022 have been collected using vehicle-mounted Smartphones (Cameras in the case of Norway). Thus, the models trained using the RDD2022 data would be capable of damage detection for data collected through moving vehicles, providing support for a quick inspection of a large area.
v. Data challenges can be organized using RDD2022. For instance, the Crowd Sensing-based Road Damage Detection Challenge (CRDDC’2022) is based on RDD2022. CRDDC aims to find automatic road damage detection solutions for the six underlying countries of RDD2022. Likewise, the Global Road Damage Detection Challenge (GRDDC'2020), organized as an IEEE Big Data Cup in 2020, utilized the dataset RDD2020, a part of RDD2022, to assess the solutions proposed by participants for road damage detection in India, Japan, and the Czech Republic[6,8].
vi. Machine learning researchers may utilize the RDD2022 data to benchmark the performance of different algorithms targeting similar applications, such as image classification, object detection, etc.

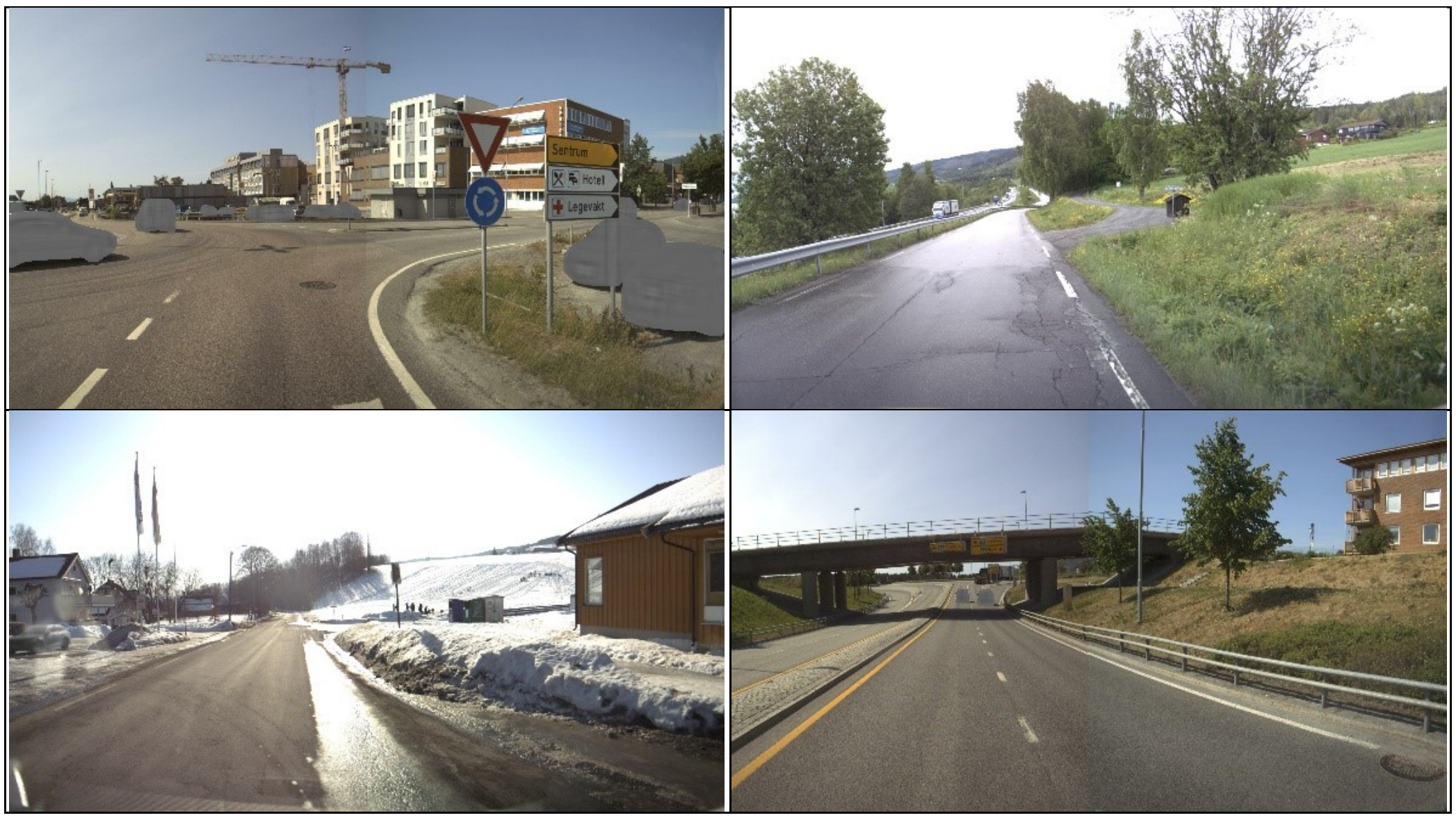

Figure 20: Sample images from Norway

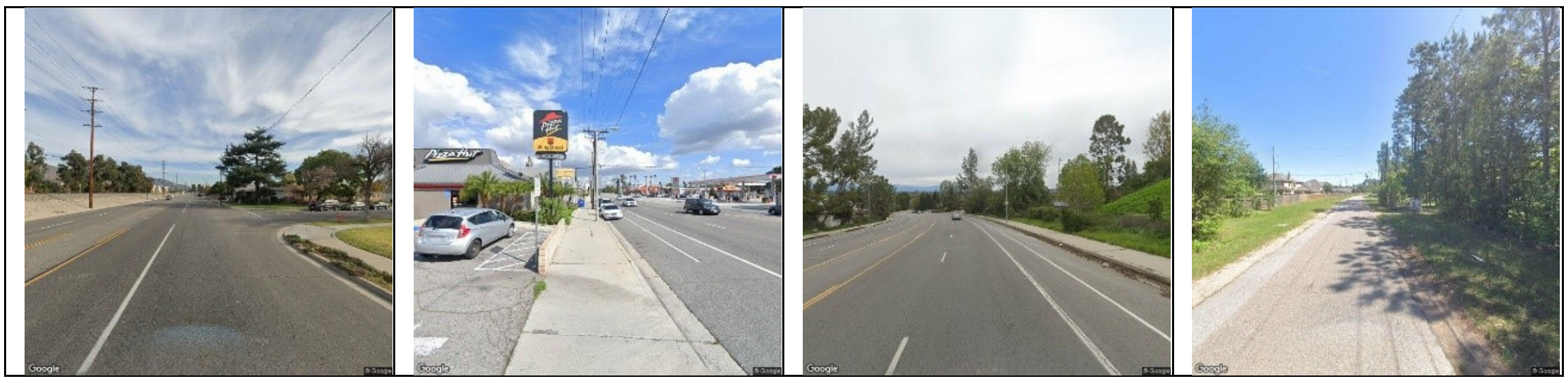

Figure 21: Sample images from the United States of America

b. With Augmentation
    i. By adding new images: Users may refer to image acquisition in the Methods section of the manuscript and may augment the RDD2022 with new images targeting multiple applications such as:
        - Covering more countries: Adding images from other countries would help automate road damage detection for those countries and improve the performance for the currently considered six countries.
        - More images for the underlying six countries: Currently, the RDD2022 dataset captures a massive number of images compared to other datasets. However, it is still not exhaustive in capturing all possible conditions (climatic/geographical) of the underlying six countries. New images may be collected to represent the heterogeneous scenarios in these countries better and train more robust models.
        - Covering new road types: Currently, the images in RDD2022 mainly capture the flexible type of roads. Application for rigid pavement may be considered by augmenting RDD2022 with new images.
    ii. By adding new annotations (damage categories) – The current version of the data supports the detection and classification of road cracks (longitudinal, transverse, and alligator) and potholes. Users may refer to the annotation pipeline in the Methods section of the manuscript and may augment the RDD2022 to capture the information of more damage categories – such as Rutting, Raveling, etc.
        - This will widen the scope of application for monitoring road conditions. The models trained using the augmented RDD2022 would help provide more specific inputs to the road agencies for requisite maintenance of the roads captured in the images.

(ii) Only images: The users may utilize the images from RDD2022 and generate new annotations targeting new applications. For instance:
    a. If the users want to keep the domain of the application the same as RDD2022, that is, Road Damage Detection - Pixel-wise annotations may be created for segmentation-based applications.
    b. Likewise, the users may annotate the images targeting other domains such as road asset identification, traffic monitoring, congestion detection, etc. Here, it may be noted that the vehicles in the Norwegian dataset included in RDD2022 have been masked in the images. Hence, the vehicle-related applications using RDD2022 may be targeted for Japan, India, the United States, the Czech Republic, and China, but not for Norway.

## Acknowledgments

We thank the researchers Hiroshi Omata and Takehiro Kashiyama for their support in organizing the CRDDC’2022. The research is partially sponsored by the Japan Society of Civil Engineers. The authors also acknowledge the participants of CRDDC - Madhavendra Sharma, Dr. Muneer Al- Hammadi, Mamoona Birkhez Shami, Dr. Alex Klein-Paste, Dr. Helge Mork, and Dr. Frank Lindseth, from Norwegian University of Science and Technology, Norway, Dr. Van Vung Pham and Du Nguyen from Sam Houston State University, Texas, United States, and Jingtao Zhong, Hanglin Cheng, Jing Zhang from Southeast University, China, for contributing the respective datasets.